\documentclass[10pt,twocolumn,letterpaper]{article}

\usepackage{multirow}
\usepackage{booktabs}
\usepackage{amssymb}
\usepackage{amsmath}
\usepackage{graphicx}
\usepackage{colortbl}
\usepackage[table]{xcolor}
\usepackage[normalem]{ulem}

\usepackage[pagenumbers]{cvpr}   

\definecolor{cvprblue}{rgb}{0.21,0.49,0.74}
\usepackage[pagebackref,breaklinks,colorlinks,allcolors=cvprblue]{hyperref}

\def\paperID{44065}
\def\confName{CVPR}
\def\confYear{2026}

\title{EReCu: Pseudo-label Evolution Fusion and Refinement with Multi-Cue Learning for Unsupervised Camouflage Detection}

\author{
Shuo Jiang$^{1\dagger}$
Gaojia Zhang$^{1\dagger}$
Min Tan$^{1\dagger*}$
Yufei Yin$^{2\dagger}$
Gang Pan$^{3}$\\
{\small $^{1}$Zhejiang Key Laboratory of Space Information Sensing and Transmission, Hangzhou Dianzi University}\\
{\small $^{2}$Laboratory of Complex Systems Modeling and Simulation, School of Computer Science and Technology, Hangzhou Dianzi University}\\
{\small $^{3}$College of Computer Science and Technology, Zhejiang University}\\
{\tt\small jiangshuo@hdu.edu.cn, tanmin@hdu.edu.cn}
}

\begin{document}

\maketitle

\begingroup
\renewcommand\thefootnote{}
\footnotetext{%
\begin{minipage}[t]{\columnwidth}
\raggedright
$^{\dagger}$ Equal contribution.\\
$^{*}$ Corresponding author.
\end{minipage}}
\endgroup

\begin{abstract}

Unsupervised Camouflaged Object Detection (UCOD) remains a challenging task due to the high intrinsic similarity between target objects and their surroundings, as well as the reliance on noisy pseudo-labels that hinder fine-grained texture learning. While existing refinement strategies aim to alleviate label noise, they often overlook intrinsic perceptual cues, leading to boundary overflow and structural ambiguity. In contrast, learning without pseudo-label guidance yields coarse features with significant detail loss. To address these issues, we propose a unified UCOD framework that enhances both the reliability of pseudo-labels and the fidelity of features. Our approach introduces the Multi-Cue Native Perception module, which extracts intrinsic visual priors by integrating low-level texture cues with mid-level semantics, enabling precise alignment between masks and native object information. Additionally, Pseudo-Label Evolution Fusion intelligently refines labels through teacher-student interaction and utilizes depthwise separable convolution for efficient semantic denoising. It also incorporates Spectral Tensor Attention Fusion to effectively balance semantic and structural information through compact spectral aggregation across multi-layer attention maps. Finally, Local Pseudo-Label Refinement plays a pivotal role in local detail optimization by leveraging attention diversity to restore fine textures and enhance boundary fidelity. Extensive experiments on multiple UCOD datasets demonstrate that our method achieves state-of-the-art performance, characterized by superior detail perception, robust boundary alignment, and strong generalization under complex camouflage scenarios. Code is available at \href{https://github.com/JSLiam94/EReCu}{https://github.com/JSLiam94/EReCu}.
\end{abstract}

\begin{figure}[htbp]
    \centering
    \includegraphics[width=\columnwidth]{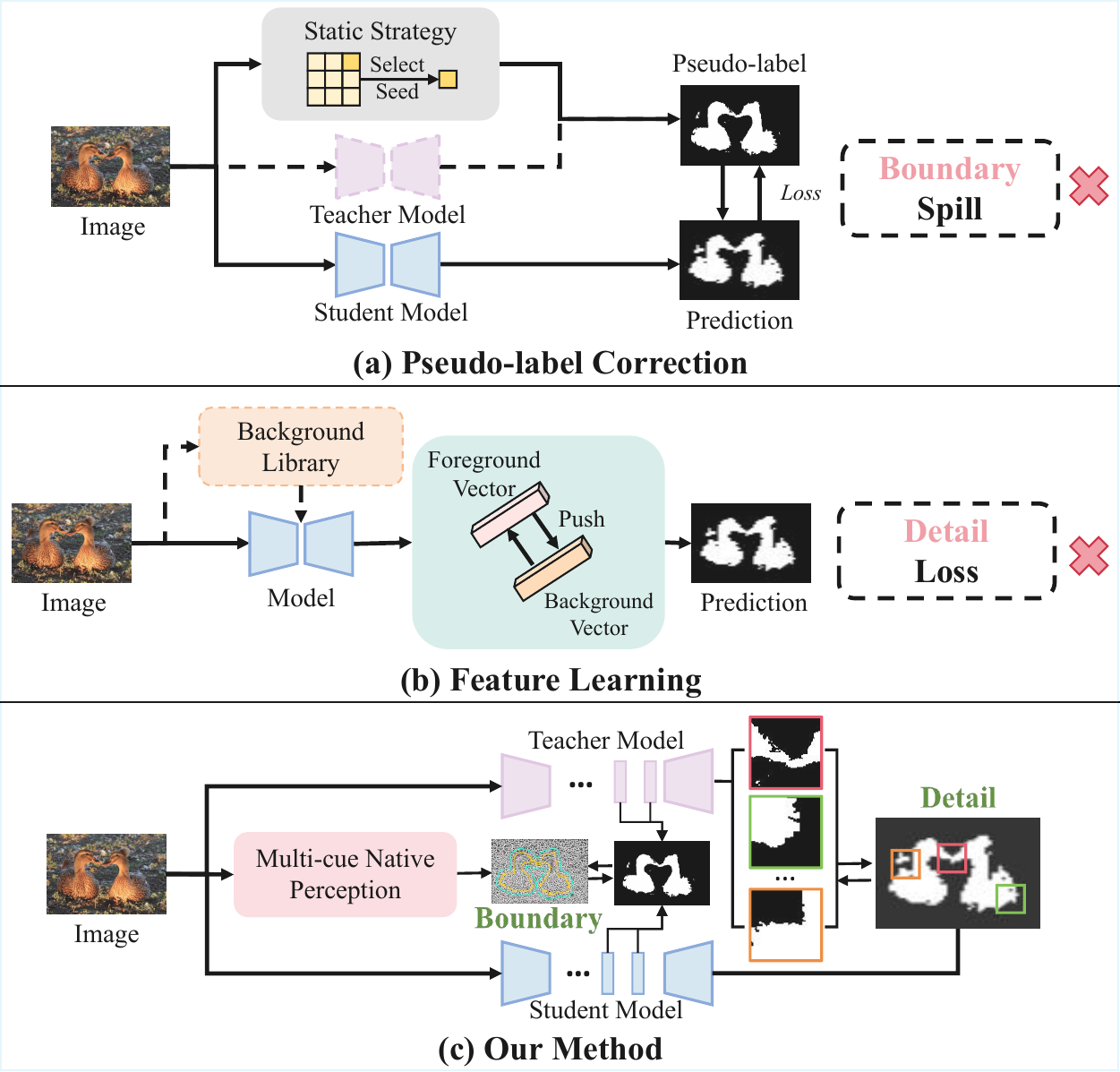} 
    \caption{UCOD paradigm comparison. 
    Traditional pseudo-label correction (a) suffers from boundary overflow due to the lack of native image cues, while feature learning methods (b) tend to generate blurred details due to the absence of pseudo-labels. Our method (c) combines pseudo-label guidance with multi-cue perception, yielding sharper boundaries and richer details.}
    \label{fig: FANS}
\end{figure} 

\section{Introduction}

Camouflage is a natural strategy that enables organisms to conceal themselves within visually complex environments~\cite{Xiao2024}. This biological principle motivates camouflaged object detection (COD), a challenging yet impactful task with applications in ecological monitoring and intelligent perception systems, as discussed in recent studies~\cite{pang2024zoomnext,hao2025simple}. However, camouflaged objects often exhibit weak texture contrast and tightly entangled boundaries with their surroundings, which undermines conventional saliency-driven detectors~\cite{zhang2025frequency,fan2023advances}.

Despite steady advances in fully supervised COD~\cite{zhang2025hunt,wang2025polarization,luo2024vscode}, existing methods still depend on dense, pixel-level masks that are costly to obtain and inherently ambiguous~\cite{HAN2025113993,sun2025conditional,Ruan_Yuan_Lin_Liao_Li_Xiong_Liu_Fu_2025}. The annotation burden limits the dataset's scale and ecological diversity, hindering the model's generalization under real-world conditions~\cite{zhao2023nowhere,he2023strategic}. This motivates Unsupervised Camouflaged Object Detection (UCOD), which learns COD without manual labels.

Existing UCOD frameworks mainly follow two paradigms: pseudo-label-guided and feature-learning-based, each with inherent bottlenecks. Early pseudo-label methods relied on static generation (e.g., background seeding), yielding fixed labels from pixel similarity to predefined background seeds~\cite{zhang2023unsupervised}. Such static supervision struggles to adapt to complex environments and often introduces background noise. Later teacher–student pipelines refined pseudo-labels by fusing teacher predictions with static labels~\cite{Yan_2025_CVPR,tan2023electromagnetic}, as illustrated in~\cref{fig: FANS}(a), but they rely heavily on high-dimensional embeddings while neglecting native perceptual cues, leading to inaccurate supervision and boundary overflow.
In parallel, as illustrated in~\cref{fig: FANS}(b), feature-learning approaches aim to disentangle foreground and background features through attention mechanisms or exploit background consistency~\cite{shou2025sdalsnet,du2025beyond,tan2025scatdiff}. Some studies further introduced environmental priors, such as an ecological prototype library, to strengthen background awareness~\cite{du2025shift,tan2019image,Liu2025Causal}. While these designs capture contrastive cues, they lack refinement mechanisms, causing blurred boundaries and lost details.

To overcome these limitations, we reconceptualize camouflage perception through the lens of semantic–perceptual integration. \textit{Our central insight is that semantic reliability and texture fidelity should not be optimized in isolation, but rather co-evolve via a mutual feedback loop.}
Accordingly, we introduce a self-evolving teacher–student framework in which native perceptual cues continuously guide the evolution of pseudo-labels, while perceptual learning simultaneously benefits from progressively denoised supervision.
This tightly coupled design enables the model to enhance both semantic coherence and structural precision simultaneously, effectively bridging the long-standing semantic and perceptual gap while addressing two persistent challenges in UCOD: pseudo-label drift and detail degradation.

Specifically, built upon a teacher–student architecture based on DINO~\cite{caron2021emerging}, our method realizes this co-evolution through three synergistic designs:  (1) \textit{Multi-Cue Native Perception (MNP)}, which enforces alignment between masks and intrinsic image patterns; (2) \textit{Pseudo-Label Evolution Fusion (PEF)}, which models pseudo-label evolution and denoising patterns across layers using efficient depthwise-separable convolutions and spectral fusion; and (3) \textit{Local Pseudo-Label Refinement (LPR)}, which preserves high-confidence object details from the selected teacher’s attention maps to construct accurate local pseudo-labels.

This cooperative mechanism allows the student network to iteratively denoise, localize, and refine camouflaged regions, producing structure-preserving object masks without manual annotation. Our main contributions are as follows:
i) We present a unified UCOD framework that integrates pseudo-label evolution with native perceptual learning via a self-evolving teacher–student mechanism.
ii) We design three complementary modules: MNP, PEF, and LPR, which jointly facilitate hierarchical semantic refinement, strengthen local structures, and improve texture-aware perception.
iii) We conduct comprehensive evaluations on multiple UCOD benchmarks, demonstrating the effectiveness of our EReCu framework.

\section{Related Work}

\begin{figure*}[tbp] 
\centering
\includegraphics[width=\textwidth]{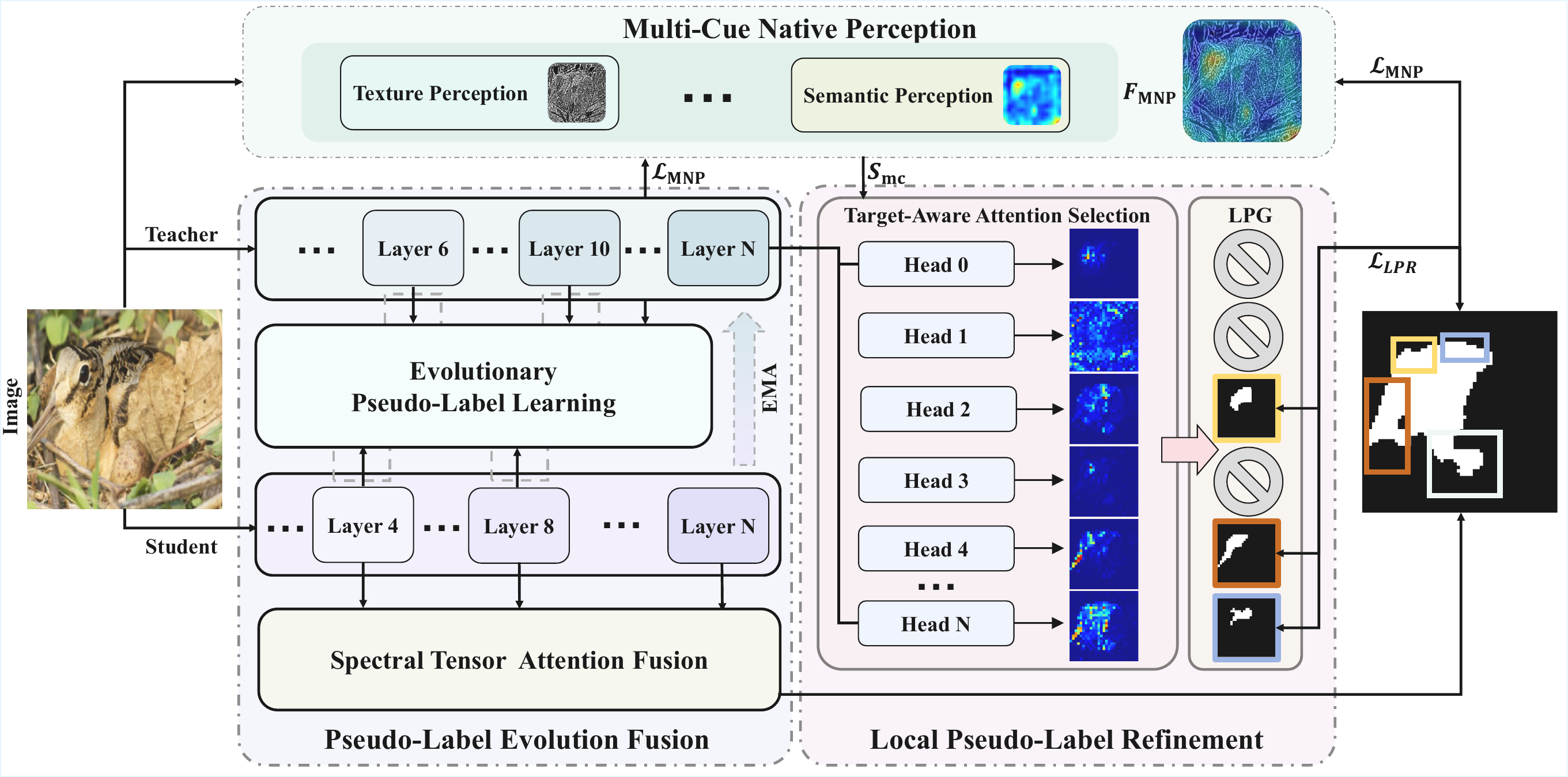} 
\caption{The proposed EReCu adopts a DINO-based teacher–student architecture. The MNP module captures texture cues from the input to refine pseudo-labels and maintain accurate object boundaries. The EPL module enables students to learn robust semantic representations by leveraging teacher deep features in shallow layers. The STAF module collects multi-layer attention maps to create low-noise masks. Lastly, LPG generates local pseudo-labels from high-confidence areas of TAS-selected maps, refining boundary fidelity.}
\label{fig: model}
\end{figure*}

\subsection{Unsupervised Camouflaged Object Detection}

UCOD aims to segment hidden objects without manual annotations, facing two fundamental challenges: the absence of supervision and the extremely low contrast between foreground and background. Pseudo-label refinement approaches, such as UCOS-DA~\cite{zhang2023unsupervised}, leverage static pseudo-labels generated by DINO~\cite{caron2021emerging} for self-training. However, the inherent label noise often induces contour overflow and semantic drift. UCOD-DPL~\cite{Yan_2025_CVPR} addresses this issue by fusing teacher-model predictions with fixed strategies to produce dynamic pseudo-labels, thereby reducing noise. However, the lack of native image information guidance still results in contour overextension. In contrast, feature-learning-based strategies eliminate reliance on pseudo-labels. SdalsNet~\cite{shou2025sdalsnet} employs self-distilled attention shifts to disentangle foreground and background feature vectors, followed by multi-stage refinement. EASE~\cite{du2025shift} introduces environment prototype retrieval to reverse-model the background, enhancing the salience of camouflaged targets. While these methods more effectively constrain target contours, the lack of explicit pseudo-label supervision often results in somewhat blurred fine-grained object structures.

\subsection{Unsupervised Object Segmentation}

Unsupervised Object Segmentation (UOS) aims to learn segmentation masks without human annotations. Early DINO~\cite{caron2021emerging} models demonstrated latent object discovery capabilities, inspiring feature-driven localization methods. LOST~\cite{2021LOST} constructs a global similarity graph from Transformer key features and expands object regions through graph-theoretic seed growth; however, its reliance on specific feature types and seed heuristics limits generalization. FOUND~\cite{2023FOUND} adopts a background-first paradigm, selecting background seeds through attention maps to produce coarse background masks, allowing foreground inference by contrast. TokenCut~\cite{wang2022self} builds a full token-affinity graph to exploit spatial/structural cues and uses Normalized Cut for foreground and background partitioning; although it outperforms LOST in coverage and consistency, its boundaries remain coarse. CutLER~\cite{2023CutLER} builds on TokenCut via MaskCut, iteratively producing candidate masks and using them for self-training, thus extending discovery to unsupervised detection and instance segmentation. U2Seg~\cite{2024u2seg} integrates Self-supervised learning features with clustering-derived pseudo-labels, while SLMP~\cite{2025SLMP} leverages Sparsemax to assign point features to object parts, enabling structured shape abstraction and explicit reconstruction.

Although existing UOS models excel at general visual representation learning, directly transferring them to COD is highly challenging: the inherently low foreground–background discriminability hampers contrastive/reconstruction features from reliably capturing accurate boundaries and fine-grained semantics.




\section{The Proposed Method: EReCu}

\subsection{Overall Framework}
To address two significant limitations of existing UCOD approaches, we propose \textbf{EReCu}, a unified teacher--student framework that integrates multi-cue native perception with evolutionary pseudo-label refinement to enable structure-preserving camouflage object detection. As illustrated in~\cref{fig: model}, our framework forms a cohesive pipeline with three key components: (1) \textbf{MNP}, which extracts native texture and semantic cues to provide reliable supervision signals; (2) \textbf{PEF}, which leverages these cues to evolve semantically stable global pseudo-labels through teacher--student interaction and spectral tensor attention fusion; and (3) \textbf{LPR}, which utilizes native cues to generate target-aware local pseudo-labels for refining details overlooked by global predictions. Specifically, the teacher branch provides stable semantic guidance, while the student branch progressively learns refined camouflage masks under evolving pseudo-label supervision. In particular, MNP provides $F_{\text{MNP}}$ and $S_{\text{mc}}$ to regularize pseudo-label evolution in PEF and guide reliable attention selection in LPR. PEF first produces global pseudo-labels, which are then refined by LPR to recover boundary and structural details. This coordinated design enables native perceptual cues to guide both global evolution and local refinement, thereby enabling robust and accurate camouflage object detection.

\subsection{Multi-Cue Native Perception (MNP)}

\begin{figure}[tbp]
    \centering
    \includegraphics[width=0.95\columnwidth]{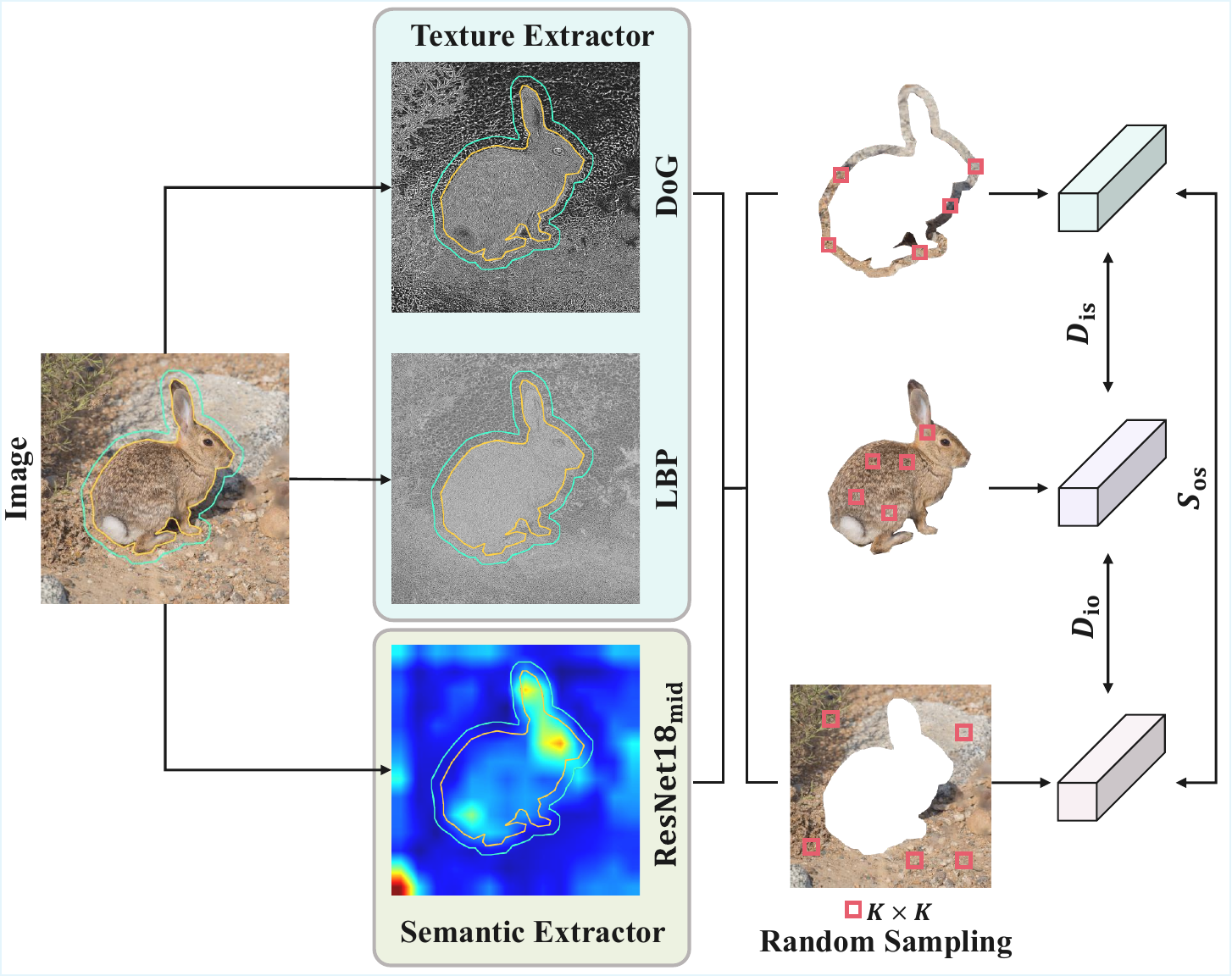} 
\caption{
Overview of the proposed MNP. It extracts native perceptual cues by combining low-level texture features (LBP, DoG) with mid-level semantics (frozen ResNet-18), and uses random sampling to ensure robust multi-cue similarity estimation.
}
    \label{fig: MNP}
\end{figure} 

The MNP module serves as the cornerstone of our multi-cue framework, providing native perceptual guidance for both pseudo-label evolution in PEF and local refinement in LPR. Although camouflage arises from high visual similarity to the surroundings, subtle yet discriminative texture variations persist within the raw image. By extracting these native low-level cues, MNP enables more reliable camouflage decoding while preserving structural fidelity.
To this end, we construct a multi-cue native representation $F_{\text{MNP}}$ together with a quality metric $S_{\text{mc}}$ that jointly provide stable, image-driven supervision. These cues anchor the pseudo-label evolution in PEF and the local corrections in LPR, ensuring that all refinements remain tightly aligned with the intrinsic characteristics of the original image.

\subsubsection{Multi-Cue Representation}
First, by employing a comprehensive group of descriptors $\{F_{l}^{i}\}$ ($i=1,\dots, T$) that represent local edge and texture variations, we obtain a robust texture feature vector:
\begin{equation}
\begin{aligned}
F_{\text{text}} &= \mathcal{C}\big( F_{l}^{1}(I), ..., F_{l}^{T}(I) \big), \qquad
F_{\text{sem}} = F_r(I).
\end{aligned}
\label{eq:text_sem}
\end{equation}
wherein $\mathcal{C}$ denotes the concatenation operator, $T$ is the number of texture descriptors, $F_l^i(\cdot)$ denotes the $i$-th descriptor extractor, and $F_r(\cdot)$ denotes the intermediate feature extractor instantiated by a frozen ResNet-18.
Consequently, we obtain the final multi-cue feature \( F_{\text{MNP}} \):
\begin{equation}
F_{\text{MNP}} = \mathcal{C}(F_{\text{text}}, F_{\text{sem}}).
\label{eq:concat}
\end{equation}

\begin{figure}[tbp]
    \centering
    \includegraphics[width=\columnwidth]{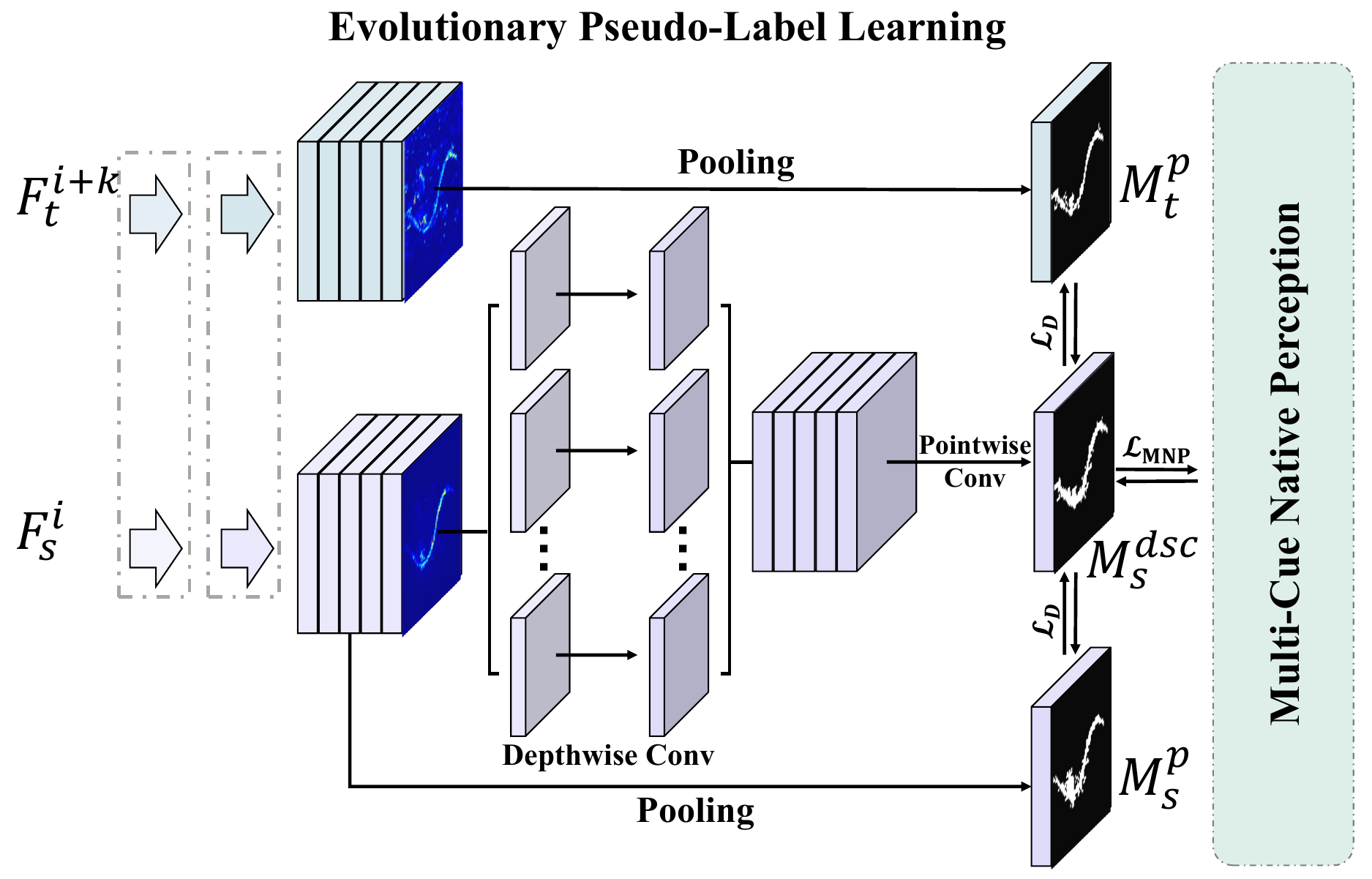} 
\caption{Overview of EPL module. 
It enables interaction between shallow student features $F_s^i$ and deep teacher features $F_t^{i+k}$ via DSC, and progressively refines pseudo-masks $M_s^{\mathrm{dsc}}$ and $M_t^{\mathrm{p}}$ using a hierarchical loss combining Dice and perceptual terms.
}
    \label{fig: EPL}
\end{figure}

\subsubsection{Multi-Cue Quality Metric}

To quantify the fore–background separability encoded in $F_{\text{MNP}}$, we introduce a multi-cue metric $S_{\text{mc}} $ together with its associated loss $\mathcal{L}_{\text{MNP}}$. Given a predicted mask or pseudo-label denoted as $M$, we partition the image into three structurally meaningful regions using a threshold $\tau_R$:
\begin{equation}
\begin{aligned}
R_{\text{i}} &= \{ p \mid M(p) > \tau_R \}, \\
R_{\text{s}} &= \text{DL}(M, SE) \setminus R_{\text{i}}, \\
R_{\text{o}} &= \{ p \mid M(p) < \tau_R \} \setminus R_{\text{s}},
\label{eq:regionset}
\end{aligned}
\end{equation}
where \(\text{DL}\) denotes morphological dilation with a circular structuring element 
\(SE\).
These regions respectively capture the inside (\(R_{\text{i}}\)), boundary (\(R_{\text{s}}\)), and exterior (\(R_{\text{o}}\)) contexts of the object, enabling structured cue comparison across them.

For each region $R_{x}$, we extract its characteristic representation $\bar{F}_{x}$ from \( F_{\text{MNP}} \). We then compute three complementary cosine-based relations:
\begin{equation}
\begin{aligned}
D_{\text{io}} &= 1 - \text{CSim}(\bar{F}_{\text{i}}, \bar{F}_{\text{o}}), \\
D_{\text{is}} &= 1 - \text{CSim}(\bar{F}_{\text{i}}, \bar{F}_{\text{s}}), \\
S_{\text{so}} &= \text{CSim}(\bar{F}_{\text{s}}, \bar{F}_{\text{o}}).
\end{aligned}
\label{eq:D_AND_S}
\end{equation}

Herein, \(D_{\text{io}}\), \(D_{\text{is}}\), and \(S_{\text{so}}\) effectively measure interior–exterior separation, interior–boundary contrast,  and boundary-exterior similarity, respectively.

As depicted in~\cref{fig: MNP}, to address the challenges posed by irregular differences in shape across regions, the corrected cosine similarity is computed by randomly sampling patches of size $K \times K$ over N rounds:
\begin{equation}
\text{CSim}(A, B) = \frac{1}{N} \sum_{m=1}^{N} 
\frac{\langle \bar{A}_m, \bar{B}_m \rangle}{\| \bar{A}_m \| \| \bar{B}_m \|}.
\label{eq:Csim}
\end{equation}

Here, $A$ and $B$ correspond to the patch-level representations 
$\bar{F}_{x}$ defined in~\cref{eq:D_AND_S}. This correction yields stable similarity estimation regardless of region size. Finally, the multi-cue metric $S_{\text{mc}}$ and loss $\mathcal{L}_{\text{MNP}}$ are defined as:
\begin{equation}
S_{\text{mc}} = \frac{D_{\text{io}} + D_{\text{is}} + S_{\text{so}}}{3}, \quad \mathcal{L}_{\text{MNP}} = 1 - S_{\text{mc}}.
\label{eq:Smc}
\end{equation}
A higher $S_{\text{mc}}$ indicates stronger fore–background separation, providing reliable native-cue signals to pseudo-labels. In the following modules, $\mathcal{L}_{\text{MNP}}$ is consistently computed on the regions induced by the current candidate mask, so that pseudo-label updates remain aligned with the native image cues encoded in $F_{\text{MNP}}$.

\subsection{Pseudo-Label Evolution Fusion (PEF)}

PEF integrates two complementary components, Evolutionary Pseudo-Label Learning (EPL) and Spectral Tensor Attention Fusion (STAF), to provide reliable global supervision for camouflage detection. Specifically, EPL generates coarse yet semantically reliable pseudo-labels through teacher--student co-evolution under native-cue regularization, progressively stabilizing the global supervisory signal. Building on this, STAF further fuses multi-level student-attention cues into a unified representation that preserves semantic structure and fine-grained details, yielding a compact fused prediction. Together, these two components couple global pseudo-label evolution with multi-level attention fusion, yielding stable, discriminative, and detail-aware global guidance for the subsequent local refinement in LPR.

\subsubsection{Evolutionary Pseudo-Label Learning}

Building upon MNP’s native cues, we introduce the EPL module to address the limitations of traditional UCOD pseudo-labels, which over-rely on deep semantic features and often overlook fine-grained details, resulting in oversmoothed boundaries. EPL enables shallow student layers to interact with deep teacher features while preserving structural integrity, guided by native perceptual cues. This interaction facilitates semantic denoising and iteratively refines pseudo-labels through temporal evolution. EPL fosters co-evolution between semantic abstraction and detail recovery, yielding more accurate camouflage masks.

Let \(F_s^i\) represent the shallow feature maps from the student branch and \(F_t^{i+k}\) denote the deep feature maps from the teacher branch. To extract task-relevant information while preserving structural integrity, we initially apply a Depthwise Separable Convolution (DSC) to the student's shallow feature map \(F_s^i\). The DSC decomposes standard convolution into depthwise and pointwise operations, substantially lowering computational cost while enabling separate refinement of spatial and channel-wise features.
This operation adaptively enhances fine textures and boundary structures, which are critical for effective camouflage detection.
Subsequently, we extract pseudo-mask candidates from both the student and teacher branches:
\begin{equation}
M_s^{\text{dsc}} = \mathcal{B}\big(\text{DSC}(F_s^i)\big), \quad
M^{\text{p}}(\cdot) = \mathcal{B}\big(\text{Pool}(\cdot)\big),
\end{equation}
where \(\text{Pool}\) denotes semantic pooling that fuses average and maximum features, and \(\mathcal{B}(\cdot)\) is for binarization that converts probability maps into mask predictions. Accordingly, the teacher branch produces a coarse pseudo-mask 
$M_t^{\text{p}} = M^{\text{p}}(F_t^{i+k})$, 
while the student branch yields 
$M_s^{\text{p}} = M^{\text{p}}(F_s^i)$.
As illustrated in~\cref{fig: EPL}, the evolutionary pseudo-label optimization process at iteration \(r\) is defined as:

\begin{equation}
\begin{aligned}
M_s^{\text{dsc}(r+1)} = \arg \min_{M_s^{\text{dsc}}} \Big[
   & \mathcal{L}_{\text{D}}(M_s^{\text{dsc}(r)}, M_s^{\text{p}(r)}) \\
   & + \mathcal{L}_{\text{D}}(M_s^{\text{dsc}(r)}, M_t^{\text{p}(r)}) \\
   & + \mathcal{L}_{\text{MNP}}(M_s^{\text{dsc}(r)}, F_{\text{MNP}}^{(r)})
\Big],
\end{aligned}
\end{equation}
where both \(\mathcal{L}_{\text{D}}\) employ the Dice loss, while \(\mathcal{L}_{\text{MNP}}\) regularizes the updates of the pseudo-labels by leveraging native perceptual features \(F_{\text{MNP}}\) as described in~\cref{eq:concat,eq:Smc}.

This iterative optimization jointly exploits hierarchical features, temporal consistency, and native image cues to drive evolutionary pseudo-label learning. The resulting pseudo-labels \(M_s^{\text{dsc}}\) provide global supervision for learning the student prediction, reinforcing the training signal. Consequently, EPL progressively improves pseudo-label quality for boundary-accurate, detail-preserving camouflage segmentation, with the evolution governed by both teacher–student agreement and native-cue regularization rather than by semantic guidance alone.

\subsubsection{Spectral Tensor Attention Fusion}

\begin{figure*}[tbp]
    \centering
    \includegraphics[width=\textwidth]{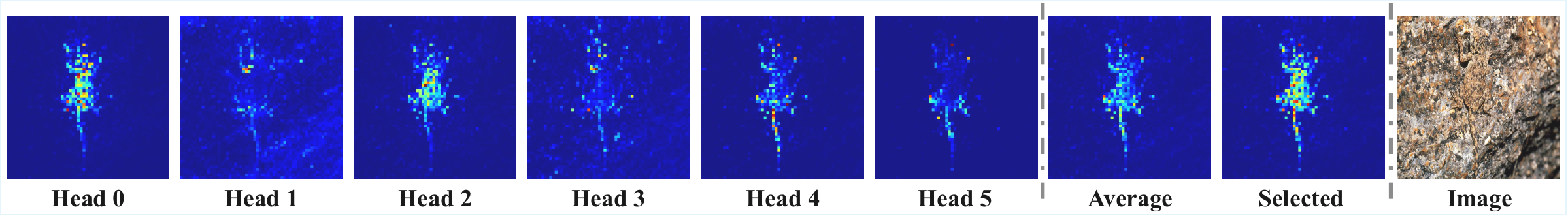}
\caption{
Visualization of MHSA reveals that different heads focus on distinct visual cues. Comparing individual heads, their average, and an attention-selected aggregation against the original image shows that the proposed attention selection exhibits low attention entropy and conforms to intrinsic image characteristics, thereby reducing noise interference while preserving details.
}
    \label{fig: attn}
\end{figure*}

To effectively integrate multi-layer student attention maps, we introduce the STAF module. Unlike naive weighted aggregation, which may cause semantic dilution and structural loss, STAF combines low-rank Tucker decomposition with Singular Value Decomposition (SVD)-based spectral filtering to fuse high-dimensional features while preserving informative components and suppressing noise. This process produces a consolidated prediction $M^{\mathrm{fu}}_s$ that maintains semantic and structural consistency and provides iterative reinforcement to the student network.

Specifically, we extract three student attention maps from hierarchical levels $\frac{\text{Len}}{3}$, $\frac{2\text{Len}}{3}$, and the final layer. These maps are stacked into a third-order tensor representation $\mathcal{T}_s \in \mathbb{R}^{3 \times C \times HW}$, where $C$ denotes the channel number and $H$ and $W$ denote spatial height and width. A Tucker decomposition is then applied to capture correlations across levels, channels, and spatial locations:
\begin{equation}
\mathcal{T}_s \approx \mathcal{G} \times_1 U^{(1)} \times_2 U^{(2)} \times_3 U^{(3)},
\end{equation}
where $\mathcal{G}$ is the core tensor, $\{U^{(i)}\}_{i=1}^3$ are factor matrices describing principal subspaces along each mode, and $\times_x$ denotes tensor--matrix multiplication along the $x$-th mode. This decomposition compactly captures the shared structure across hierarchical attention levels while reducing redundancy in the original tensor.

To extract the dominant spectral components, the compact core tensor $\mathcal{G}$ is first unfolded into a matrix form, making its cross-dimensional correlations more amenable to spectral analysis.We then apply truncated SVD to $\mathcal{U}(\mathcal{G})$ to obtain a rank-$t$ approximation:
\begin{equation}
A_s^{\text{fu}} = P_t \Sigma_t Q_t^\top \approx \mathcal{U}(\mathcal{G}),
\end{equation}
where $\mathcal{U}(\cdot)$ denotes tensor unfolding, and $P_t$, $\Sigma_t$, and $Q_t$ retain the top $t$ singular components. Thus, the fused representation preserves dominant spectral energy while filtering out low-energy noise and unstable responses, providing a more robust and reliable basis for subsequent mask prediction. The final fused prediction $M_s^{\text{fu}}$ used in the LPR module is obtained by linearly projecting $A_s^{\text{fu}}$ to the mask space, followed by a sigmoid activation:
\begin{equation}
M_s^{\text{fu}} = \text{Sigmoid}(W A_s^{\text{fu}} + b),
\label{eq:Msfu}
\end{equation}
where $W$ and $b$ denote projection weights and bias. Here, “spectral” refers to retaining dominant singular-spectrum energy under low-rank approximation, enabling efficient fusion with complexity $\mathcal{O}(r^2 d)$, where $r$ is the retained rank and $d$ is the unfolded feature dimension, with $r \ll d$.

\subsection{Local Pseudo-Label Refinement (LPR)}

While global pseudo-labels effectively capture central object regions, they often miss boundary and texture details. To address this issue, we developed the LPR module. As shown in~\cref{fig: attn}, attention maps highlight high-confidence target regions with rich structural cues, while different heads emphasize distinct regions of interest. Leveraging this property, LPR exploits the spatial diversity of DINO's multi-head self-attention (MHSA) to refine pseudo-labels locally. It consists of two components: Target-Aware Attention Selection (TAS), which selects target-focused attention heads using MNP's perceptual cues and attention entropy, and Local Pseudo-Label Generation (LPG), which uses the selected maps to generate fine-grained local pseudo-labels that preserve semantic coherence and structural consistency.

\subsubsection{Target-aware Attention Selection}

Given multiple attention heads \(\{A_k\}\) obtained from the MHSA layer, we propose a TAS mechanism to enhance the model's focus on relevant features. The first step in TAS involves measuring the attention concentration of each head by calculating its focusing entropy:

\begin{equation}
E_k = \frac{- \sum_p A_k(p) \log A_k(p)}{\log n},
\end{equation}
where \(n\) denotes the total number of elements in the attention map. A lower entropy value \(E_k\) generally indicates a more concentrated attention distribution, reflecting a head's effective focusing capability.

\begin{figure*}[tbp]
    \centering
    \includegraphics[width=2\columnwidth]{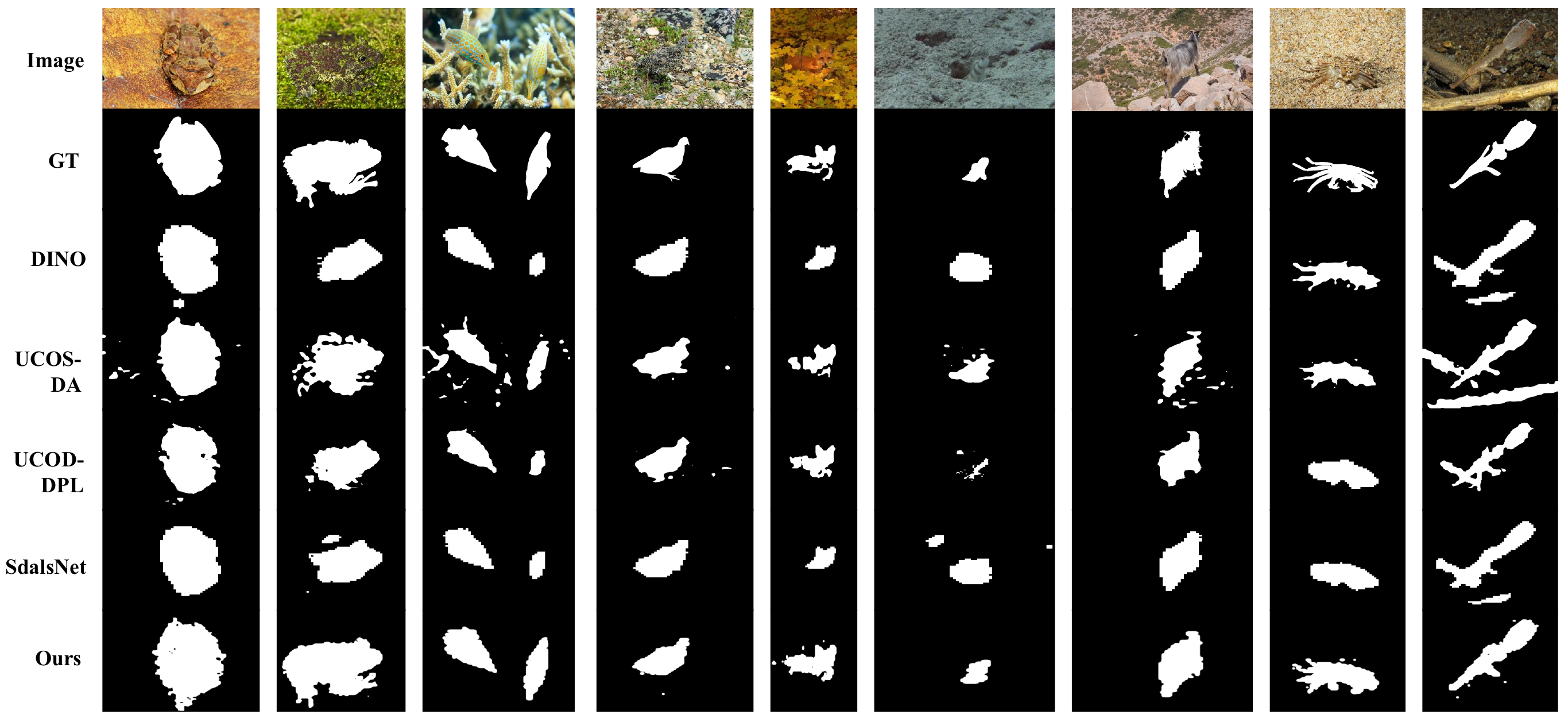} 
\caption{
Visual comparison of EReCu with existing methods in challenging scenarios, demonstrating that our method achieves clearer and more accurate segmentation boundaries while effectively detecting objects with depth-induced artifacts. GT is for ground-truth.
}
    \label{fig: vis}
\end{figure*} 

To ensure semantic reliability, we further evaluate whether each attention head is consistent with the native perceptual cues in \(F_{\text{MNP}}\). Specifically, each attention map \(A_k\) is first converted into a binary response map \(\hat{A}_k=\mathcal{B}(A_k)\), on which the multi-cue metric \(S_{\text{mc}}\) in~\cref{eq:Smc} is computed. The selection criterion is then formulated as follows:
\begin{equation}
\mathcal{A}_{\text{sel}} = \{A_k \mid E_k < \tau_e \; \wedge \; S_{\text{mc}}(\hat{A}_k, F_{\text{MNP}}) > \tau_s \},
\label{eq:Asel}
\end{equation}
where \(\tau_e\) and \(\tau_s\) are learnable threshold parameters initialized at 0.5. These thresholds ensure that only the selected attention heads maintain semantic consistency. Consequently, the set \(\mathcal{A}_{\text{sel}}\) contains target-aware attention maps derived from the final layer of the teacher model, providing valuable cues regarding object structure.

\subsubsection{Local Pseudo-label Generation}
For each selected head \( A_k \) in $\mathcal{A}_{\text{sel}}$ with~\cref{eq:Asel}, a local pseudo-label \( P_k \) is dynamically generated based on the high-confidence regions of its feature map:
\begin{equation}
P_k = \mathbf{1}_{A_k(p) > \tau_k}, \quad \tau_k = \mu_{A_k} + \alpha \cdot \sigma_{A_k},
\end{equation}
where \(\mu_{A_k}\) denotes the mean of \(A_k\), \(\sigma_{A_k}\) represents its standard deviation, and \(\alpha > 1\) is a learnable parameter that adjusts the confidence level, so that only highly activated and structurally informative regions are retained as local cues.

These local pseudo-labels are used to guide the optimization of the fused student prediction \( M_s^{\text{fu}} \) in~\cref{eq:Msfu} through a hybrid loss combining Dice and Cross-Entropy terms:
\begin{equation}
\mathcal{L}_{\text{LPR}} = \mathcal{L}_{\text{D}}\left(M_s^{\text{fu}}, \bigcup_k P_k\right) + \mathcal{L}_{\text{CE}}\left(M_s^{\text{fu}}, \bigcup_k P_k\right),
\end{equation}
where \(\mathcal{L}_{\text{D}}\) is the Dice loss that emphasizes the overlap between the student model's fused probability map \( M_s^{\text{fu}} \) 
and the union of the binary pseudo-labels \( \bigcup_k P_k \), while \(\mathcal{L}_{\text{CE}}\) is the Cross-Entropy loss. This formulation allows the student network to be iteratively corrected and refined by the high-confidence local cues, thereby promoting coherent and detail-preserving predictions.

\section{Experiments}

\subsection{Experimental Setup}

\noindent\textbf{{Used Dataset.}}
To ensure a fair comparison with existing studies, we follow the standard UCOD protocol~\cite{9156837,Yan_2025_CVPR}. Specifically, we adopt a combined training set comprising 1,000 images from CAMO-Train~\cite{le2019anabranch} and 3,040 images from COD10K-Train~\cite{9156837}. In line with the unsupervised learning paradigm, no ground-truth annotations are utilized during training. For evaluation, we assess our method on four widely used COD benchmarks: CHAMELEON~\cite{Skurowski2017Chameleon} (76 images), CAMO~\cite{le2019anabranch} (250 images), COD10K~\cite{9156837} (2,026 images), and NC4K~\cite{9577641} (4,121 images).

\noindent\textbf{Evaluation Metrics.}
Following common practice, we employ four metrics for comprehensive evaluation, including the structure measure ($S_m\uparrow$)~\cite{Fan_2017_ICCV}, 
the weighted F-measure ($F_\omega^\beta\uparrow$)~\cite{6909433},
the E-measure ($E_m^{\phi}\uparrow$)~\cite{Fan2018EnhancedalignmentMF},
and the Mean Absolute Error ($M\downarrow$)~\cite{6247743}.

\noindent\textbf{Implementation Details.}
Following~\cite{2023FOUND}, we adopt the self-supervised pre-trained backbone DINO~\cite{caron2021emerging} as the encoder. We utilize the Local Binary Pattern (LBP) and Difference of Gaussian (DoG) as the texture extractors ($F_l^i$ in~\cref{eq:text_sem}), while the semantic extractor utilizes ResNet-18 ($F_r$ in~\cref{eq:text_sem}), selected through empirical validation. The teacher model is updated by an Exponential Moving Average (EMA) strategy with a momentum parameter $\eta = 0.99$. The threshold $\tau_R$ in~\cref{eq:regionset} is empirically set to $0.5$.
We train the model for 25 epochs with a batch size of 32, using the AdamW optimizer~\cite{loshchilov2017decoupled} and a cosine annealing schedule for dynamic learning rate adjustment. To enhance stability and efficiency, we adopt automatic mixed precision (AMP) during training. All experiments are conducted on NVIDIA Tesla V100-SXM2 GPUs (32 GB memory) using PyTorch 2.4 with CUDA 12.1. For competing methods, we reimplement publicly available models using their released code. For EASE~\cite{du2025shift}, which does not provide source code, we report the performance values directly from the original paper. 
All methods using DINO for feature extraction employ DINO-ViT-S/8. All experiments use the same random seed of 2026 for reproducibility. 
For additional details on the hyperparameter experiments of the extractors in~\cref{eq:concat}, and the $N$,$K$ in~\cref{eq:Csim}, refer to the supplementary materials.

\begin{table*}[t]
\centering
\caption{\small Quantitative comparison of our proposed method with existing UOS and UCOD approaches across four COD datasets. \textbf{Bold} indicates the best result in each group, and \underline{underline} denotes the second-best result.}
\renewcommand{\arraystretch}{1.1}
\setlength{\tabcolsep}{3pt}
\resizebox{\textwidth}{!}{%
{\small
\begin{tabular}{l l|cccc|cccc|cccc|cccc}
\toprule
\multirow{2.4}{*}{\textbf{Type}} & \multirow{2.4}{*}{\textbf{Methods}} 
& \multicolumn{4}{c|}{\textbf{CHAMELEON}} 
& \multicolumn{4}{c|}{\textbf{CAMO-Test}} 
& \multicolumn{4}{c|}{\textbf{COD10K-Test}} 
& \multicolumn{4}{c}{\textbf{NC4K}} \\
\cmidrule(lr){3-6} \cmidrule(lr){7-10} \cmidrule(lr){11-14} \cmidrule(lr){15-18}
& & $S_m\uparrow$ & $F_\omega^\beta\uparrow$ & $E_m^{\phi}\uparrow$ & $M\downarrow$ 
& $S_m\uparrow$ & $F_\omega^\beta\uparrow$ & $E_m^{\phi}\uparrow$ & $M\downarrow$ 
& $S_m\uparrow$ & $F_\omega^\beta\uparrow$ & $E_m^{\phi}\uparrow$ & $M\downarrow$ 
& $S_m\uparrow$ & $F_\omega^\beta\uparrow$ & $E_m^{\phi}\uparrow$ & $M\downarrow$ \\
\midrule
\multirow{6}{*}{\textbf{UOS}}
&TokenCut~\cite{wang2022self}  & .6573 & .5018 & .7351 & .1379 & .6450 & .5149 & .7237 & .1605 & .6638 & .4770 & .7539 & .1023 & .7338 & .6137 & .7919 & .1083 \\
&SelfMask~\cite{9857257}  & .6522 & .5200 & .8091 & .1311 & .6596 & .5827 & .7867 & .1542 & .6495 & .4480 & .8418 & .1091 & .7306 & .6259 & .8201 & .0999 \\
&SpectralSeg~\cite{melas2022deep} & .5838 & .4634 & .6690 & .1776 & .5983 & .4726 & .6889 & .1986 & .5938 & .3816 & .6138 & .1759 & .6973 & .5672 & .7328 & .1472 \\
&A2S-v2~\cite{zhou2023texture}    & .5832 & .4711 & .6839 & .1134 & .6128 & .5466 & .7071 & .1451 & .6428 & .4932 & .7503 & .0802 & .7131 & .6509 & .8022 & .0883 \\
&FOUND~\cite{2023FOUND}     & .7161 & .6112 & .7704 & .0892 & .6913 & .6217 & .7465 & .1373 & .6783 & .5056 & .6475 & .0841 & .7459 & .6589 & .8073 & .0886 \\
&DINO~\cite{caron2021emerging}      & .6613 & .5279 & .7714 & .1220 & .6376 & .5298 & .7431 & .1568 & .6400 & .4494 & .7035 & .1032 & .6968 & .5866 & .7826 & .1085 \\
\midrule
\multirow{5}{*}{\textbf{UCOD}}
&UCOS-DA~\cite{zhang2023unsupervised}   & .6715 & .5221 & .7504 & .1256 & .6581 & .5470 & .7343 & .1637 & .6334 & .4295 & .6760 & .1219 & .7189 & .5998 & .7831 & .1070 \\
&UCOD-DPL~\cite{Yan_2025_CVPR}  & \underline{.7287} & \underline{.6154} & \underline{.8486} & \underline{.0725} & \underline{.7013} & \textbf{.6109} & .7921 & \underline{.1082} & \underline{.7090} & \underline{.5481} & \underline{.8090} & \textbf{.0601} & \underline{.7538} & \textbf{.6674} & \underline{.8447} & \underline{.0745} \\
&SdalsNet~\cite{shou2025sdalsnet}  & .7236 & .6113 & .8347 & .0810 & .6971 & .6010 & \underline{.7994} & .1174 & .6967 & .5250 & .7797 & .0717 & .7386 & .6417 & .8243 & .0850 \\
&EASE~\cite{du2025shift}      & .6760 & .5500 & .7650 & .1050 & .6530 & .5630 & .7370 & .1660 & .6730 & .5140 & .7320 & .1090 & .7280 & .6330 & .7900 & .1080 \\
\rowcolor{gray!10}

&\textbf{EReCu}      & \textbf{.7321} & \textbf{.6187} & \textbf{.8523} & \textbf{.0716} & \textbf{.7027} & \underline{.6083} & \textbf{.8003} & \textbf{.1072} & \textbf{.7221} & \textbf{.5628} & \textbf{.8185} & \underline{.0613} & \textbf{.7583} & \underline{.6642} & \textbf{.8498} & \textbf{.0742} \\
\bottomrule
\end{tabular}
}}
\label{tab: compare}
\end{table*}

\subsection{Overall performance}
We conduct both qualitative and quantitative evaluations to assess EReCu. Additional details on hyperparameter experiments are provided in the supplementary materials.

\noindent\textbf{Qualitative Analysis.}
We present qualitative comparisons between our method and recent UCOD approaches on challenging camouflage scenes, as shown in~\cref{fig: vis}. Our method produces more accurate segmentation masks with sharper boundaries and richer structural details.

\noindent\textbf{Quantitative Analysis.}
To further demonstrate its competitiveness, in addition to recent UCOD methods, we also include adapting several representative UOS models for our task. As shown in~\cref{tab: compare}, EReCu achieves state-of-the-art performance across all four COD datasets. Notably, our method consistently outperforms all UOS baselines across all datasets and achieves superior performance over SOTA UCOD methods on most metrics. This demonstrates its ability to discover and exploit implicit supervisory signals in unlabeled data, confirming robustness and generalization.

\subsection{Ablation Studies}

To investigate the role of each proposed component, we perform ablation studies on CAMO and COD10K under identical training settings, comparing model variants composed of different module combinations. As shown in ~\cref{tab: ablation}, every variant outperforms the DINO-ViT-S/8 backbone baseline, and the whole model yields the best results across datasets and metrics. These results indicate that each component contributes positively and complementarily to the framework. Below, we provide a focused analysis of the individual effects of each module.

\noindent\textbf{Effect of MNP.} Removing MNP degrades localization on texture-suppressed regions, causing more background patterns to be mistaken for camouflaged objects. This confirms that multi-cue native perception provides essential low-level texture cues for boundary discrimination.

\noindent\textbf{Role of PEF.} Removing PEF markedly degrades both structural coherence and overall accuracy. EPL is crucial for stabilizing training, as its iterative co-evolution and detail-preserving supervision maintain consistent teacher–student guidance, reduce noise in pseudo-labels, and enhance boundary fidelity. STAF enforces global–local consistency by adaptively fusing semantic and textural cues; without it, region-wise inconsistencies emerge.

\noindent\textbf{Ability of LPR.} Omitting LPR reduces local contrast and weakens recovery of subtle structures. LPR’s head-wise perceptual compensation restores edge fidelity and recovers fine details that complement global predictions.

\noindent\textbf{Module Synergies.} Variants using only one or two modules perform notably worse than configurations with three or four, indicating strong interdependence. The MNP and EPL pairing yields the most significant improvement by harmonizing native-cue alignment with pseudo-label learning. Other pairs, such as STAF with LPR, provide moderate gains but fall short of achieving the complete model. These observations confirm that the full integration of all four modules is crucial to achieving robust and accurate UCOD performance.

\begin{table}[!tbp]
\centering
\caption{Ablation study on CAMO and COD10K datasets.}
\renewcommand{\arraystretch}{1.15}
\setlength{\tabcolsep}{3.3pt}
\small
\rowcolors{6}{white}{gray!8}

\resizebox{\columnwidth}{!}{
\begin{tabular}{c c c c | c c c | c c c}
\toprule
\multicolumn{1}{c}{\multirow{2.5}{*}{MNP}} & \multicolumn{2}{c}{\multirow{1.5}{*}{PEF}} & \multicolumn{1}{c}{\multirow{2.5}{*}{LPR}} &
\multicolumn{3}{|c|}{CAMO} &
\multicolumn{3}{c}{COD10K} \\ 
\cmidrule(lr){2-3} \cmidrule(lr){5-7} \cmidrule(lr){8-10}

 & EPL & STAF & & $S_m\uparrow$ & $F_\omega^\beta\uparrow$ & $M\downarrow$ & $S_m\uparrow$ & $F_\omega^\beta\uparrow$ & $M\downarrow$ \\
\midrule
\checkmark & \checkmark & \checkmark & \checkmark & \textbf{.7027} & \textbf{.6083} & \textbf{.1072} & \textbf{.7221} & \textbf{.5628} & \textbf{.0613} \\ 
 & \checkmark & \checkmark & \checkmark & .6887 & .5923 & .1182 & .7111 & .5478 & \underline{.0653} \\ 
\checkmark &  & \checkmark & \checkmark & .6758 & .5632 & .1239 & .7038 & .5286 & .0739 \\ 
\checkmark & \checkmark &  & \checkmark & .6815 & .5823 & .1201 & \underline{.7179} & .5398 & .0675 \\ 
\checkmark & \checkmark & \checkmark &  & \underline{.6895} & \underline{.5937} & \underline{.1156} & .7109 & \underline{.5520} & .0698 \\ 
 &  & \checkmark & \checkmark & .6532 & .5411 & .1417 & .6823 & .4928 & .0916 \\ 
 & \checkmark &  & \checkmark & .6581 & .5390 & .1375 & .6881 & .4964 & .0869 \\ 
 & \checkmark & \checkmark &  & .6523 & .5382 & .1446 & .6602 & .4620 & .0935 \\ 
\checkmark &  &  & \checkmark & .6514 & .5443 & .1424 & .6851 & .4895 & .0987 \\ 
\checkmark &  & \checkmark &  & .6570 & .5476 & .1382 & .6892 & .4968 & .0942 \\ 
\checkmark & \checkmark &  &  & .6628 & .5547 & .1329 & .6928 & .5026 & .0913 \\ 
\midrule
\multicolumn{4}{c|}{DINO-ViT-S/8} & .6376 & .5298 & .1568 & .6400 & .4494 & .1032 \\ 
\bottomrule
\end{tabular}}
\label{tab: ablation}
\end{table}

\section{Conclusion and Discussion }
We propose a novel unsupervised framework for camouflaged object detection that integrates multi-cue native perception and evolutionary pseudo-label refinement, effectively enhancing the reliability and feature fidelity of the pseudo-labels, respectively. Additionally, we introduce a local pseudo-label refinement strategy that significantly enhances the capability for capturing fine details. Extensive experiments on diverse datasets demonstrate the effectiveness of our EReCu framework, with sharper boundaries and more complete object structures.

\newpage
\section{Acknowledgments}
This work was supported by the National Natural Science Foundation of China under Grant No.~62472133 and the Zhejiang Provincial Natural Science Foundation of China under Grant No.~LQN26F020053.

{
\small
\bibliographystyle{ieeenat_fullname}
\bibliography{main}
}

\end{document}